\documentclass{article}

\usepackage{PRIMEarxiv}

\usepackage[utf8]{inputenc} 
\usepackage[T1]{fontenc}    
\usepackage{hyperref}       
\usepackage{url}            
\usepackage{booktabs}       
\usepackage{amsfonts}       
\usepackage{nicefrac}       
\usepackage{microtype}      
\usepackage{lipsum}
\usepackage{fancyhdr}       
\usepackage{graphicx}       
\usepackage{tcolorbox} 
\tcbset{promptbox/.style={
    colback=gray!10,
    colframe=black,
    boxrule=0.5pt,
    arc=2pt,
    left=5pt,
    right=5pt,
    top=5pt,
    bottom=5pt,
    fonttitle=\bfseries
}}
\graphicspath{{media/}}     

\title{Simulating Generative Social Agents via Theory-Informed Workflow Design
}

\author{
  Yuwei Yan \\
  Information Hub \\
  The Hong Kong University of Science and Technology (Guangzhou) \\
  Guangzhou \\
  \And
  Jinghua Piao, Xiaochong Lan, Chenyang Shao \\
  Department of Electronic Engineering \\
  Tsinghua University \\
  Beijing\\
   \And
  Pan Hui\footnotemark[1] \\
  Information Hub \\
  The Hong Kong University of Science and Technology (Guangzhou) \\
  Guangzhou\\
  panhui@hkust-gz.edu.cn\\
  \AND
  Yong Li\footnotemark[1]\\
  Department of Electronic Engineering \\
  Tsinghua University
  Beijing \\
  yongli07@tsinghua.edu.cn \\
}

\begin{document}
\maketitle

\begin{abstract}
Recent advances in large language models have demonstrated strong reasoning and role-playing capabilities, opening new opportunities for agent-based social simulations. However, most existing agents' implementations are scenario-tailored, without a unified framework to guide the design. This lack of a general social agent limits their ability to generalize across different social contexts and to produce consistent, realistic behaviors.
To address this challenge, we propose a theory-informed framework that provides a systematic design process for LLM-based social agents. Our framework is grounded in principles from Social Cognition Theory and introduces three key modules: motivation, action planning, and learning. These modules jointly enable agents to reason about their goals, plan coherent actions, and adapt their behavior over time, leading to more flexible and contextually appropriate responses.
Comprehensive experiments demonstrate that our theory-driven agents reproduce realistic human behavior patterns under complex conditions, achieving up to 75\% lower deviation from real-world behavioral data across multiple fidelity metrics compared to classical generative baselines. Ablation studies further show that removing motivation, planning, or learning modules increases errors by 1.5–3.2×, confirming their distinct and essential contributions to generating realistic and coherent social behaviors.
\end{abstract}

\footnotetext[1]{Corresponding Authors.}

\keywords{First keyword \and Second keyword \and More}


\section{Introduction}

\textit{``What I cannot create, I do not understand.''}~\cite{feynman2015feynman} Social simulation, which echoes Feynman's classic principle, has become an important research tool in social science. It enables researchers to recreate social processes, test theories of human behavior, and study the emergence of collective patterns in controlled, artificial environments~\cite{epstein2012generative,macal2005tutorial}. Within this domain, \textit{agent-based modeling (ABM)} has long served as a foundational method~\cite{hofman2021integrating,lazer2009computational,lazer2020computational}, providing a bottom-up mechanism to analyze how individual decisions and interactions give rise to emerging social phenomena. However, traditional ABM relies on simplified and static rules that limit its ability to capture the richness and adaptability of real human behavior.

Recent advances in large language models (LLMs) have opened new possibilities for social simulation~\cite{gao2024large,wang2024survey,xi2023rise}. By leveraging strong reasoning, contextual understanding, and natural language interaction, LLM-driven simulation methods promise to overcome the rigidity of traditional ABM rules and generate more realistic social processes. Despite these advantages, most existing works are developed for specific scenarios, such as social interaction~\cite{gao2023s}, mobility planning~\cite{shao2024chain}, economic behavior~\cite{li2024econagent}, cognition understanding~\cite{piao2025emergence}, etc. Each simulation is built ad hoc for its target setting, with specific environments, goals, and interaction protocols. This leads to fragmented models that fail to reflect the interconnected nature of real human societies.

The limitation stems from the scenario-specific design methods adopted in current LLM-based social simulations. Existing works, though inspired by the ABM paradigm, often take a direct substitution approach: replace predefined decision rules with language model prompts or instructions, while the structural workflow governing agent cognition and behavior remains unspecified. 
Such scenario-tailored designs lead to disconnected agent behaviors. They lack coherence when agents need to act across multiple domains, including mobility, social interaction, and economic decision-making. In real societies, actions in one domain often influence and constrain behaviors in others. Without capturing these interactions, current LLM-based social simulations fail to reproduce realistic, interconnected social dynamics.

Building general social agents is inherently challenging due to the complexity of human decision processes, so we draw inspiration from behavioral science to guide the design. We propose a theory-informed workflow for constructing generative social agents. Grounded in Social Cognitive Theory (SCT)~\cite{bandura2001social}, which explains human behavior as emerging from the interplay between personal cognition, actions, and social environment. Building on this foundation, we design three core modules within social agent: motivation, which shapes goal formation; action planning, which translates goals into coherent behavior; and learning, which enables adaptation over time. Our main contributions can be summarized as follows:
\begin{itemize}
    \item We introduce a theory-informed workflow for constructing generative social agents, grounded in Social Cognitive Theory~\cite{bandura2001social}, and further guided by Maslow’s hierarchy of needs~\cite{maslow1943theory}, the Theory of Planned Behavior~\cite{ajzen1991theory}, and Social Learning Theory~\cite{bandura1977social} to shape the motivation, action planning, and learning modules, respectively. 
    
    \item Comprehensive experiments demonstrate that our theory-driven social agents reproduce realistic social behavior patterns under complex conditions. Across three real-world-inspired scenarios (daily mobility, social interaction, and pandemic adaptation), our agents achieve up to 65–80\% lower deviation from real human behavioral data.

    \item Through ablation studies, we show that each module contributes uniquely and substantially to behavioral realism. Removing any module leads to a quantifiable degradation in performance (e.g., up to 10× higher errors in mobility coherence without motivation modeling), highlighting the necessity of all three components for constructing socially aligned LLM agents.
\end{itemize}

\begin{figure*}[th]
  \centering
  \includegraphics[width=0.85\linewidth]{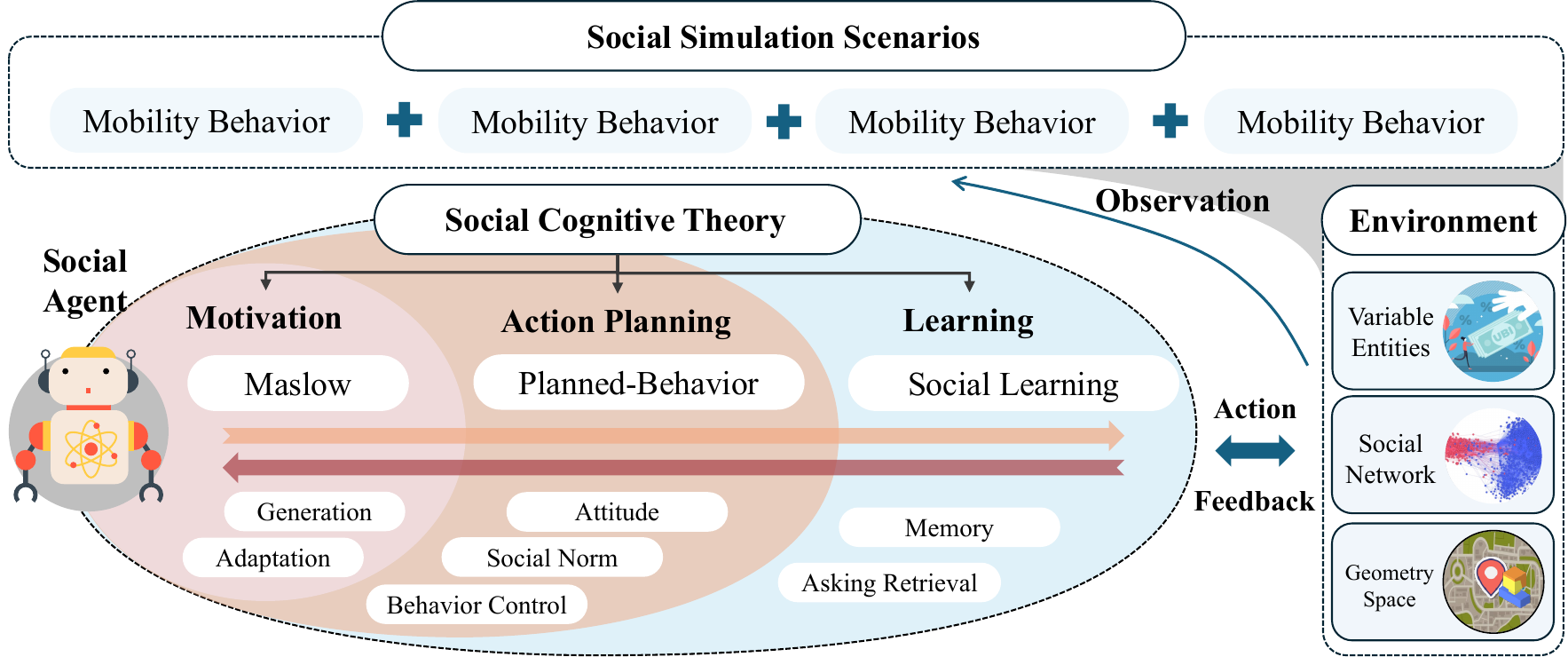}
  \caption{Generative social agent: A Theory-driven workflow.}
  \label{fig:framework}
\end{figure*}

\section{Related Work}
\subsection{Social simulation}

Social simulation is a key method in computational social science, providing a means to study how individual decisions and interactions give rise to collective patterns~\cite{macal2005tutorial, epstein2012generative}. 
Agent-based modeling has long served as a cornerstone of this field~\cite{hofman2021integrating}, representing individuals as agents that follow predefined decision rules. This approach has advanced understanding in domains such as economics, epidemiology, and urban dynamics, yet its reliance on simplified interaction rules limits its ability to capture the adaptability and diversity of real human behavior~\cite{lazer2009computational}.

Recent advances in large language models (LLMs) have opened new possibilities for social simulation. With strong reasoning, communication, and role-playing capabilities, LLMs offer a means to move beyond rigid rule-based behaviors toward more flexible, context-sensitive simulations~\cite{gao2024large,wang2024survey}. Early studies have applied LLMs to social simulations in specific domains~\cite{gao2023s, shao2024chain, li2024econagent, piao2025emergence}, demonstrating richer individual-level reasoning and more realistic interaction patterns compared to traditional ABM. Nevertheless, these efforts are often tailored to specific scenarios, where agents are designed to meet the requirements of a single task or interaction requirement. This scenario-specific approach fragments the representation of human behavior across separate models, limiting the application of LLM agents for producing realistic behaviors. 

\subsection{LLM-based agents}

The effectiveness of LLM-based social simulation highly depends on the design of their agents, as they determine how individual behaviors and interactions are modeled~\cite{hofman2021integrating,lazer2020computational}. 
While large language models (LLMs) have enhanced the capability of agents in this field, achieving promising results across diverse domains~\cite{gao2023s,shao2024chain,li2024econagent,piao2025emergence}, most of the work remains scenario-specific. This design can be effective for the target scenario, but lacks transferability to other settings and prevents agents from exhibiting or being evaluated on broader behavioral dimensions.
Although some works~\cite{park2023generative, park2024generative} design agents capable of performing actions across multiple behavioral dimensions, their implementations are often built on handcrafted prompts and intuitive designs, making it difficult to suit in a general social simulation context.
These limitations stem from the inherently scenario-tailored design choices adopted in current LLM-based agents, highlighting the need for a more general workflow for building social agents and supporting consistent behavior across simulation contexts.
\section{Generative Social Agent}
\label{sec:gsa}

\subsection{Foundations of a social agent}

As discussed earlier, existing LLM-based social agents are often tailored to isolated domains, such as mobility or social interaction, with little guidance from a shared foundation. The lack of a unified framework limits their ability to generalize across scenarios and reproduce the multifaceted nature of human behavior. To address this challenge, we propose a theory-informed approach to constructing generative social agents, illustrated in Figure~\ref{fig:framework}, which provides a coherent basis for reasoning, adaptation, and integration in diverse social scenarios.

Our design is grounded in \textbf{Social Cognitive Theory (SCT)}~\cite{bandura2001social}, which views human behavior as the outcome of continuous and reciprocal interactions among personal cognition, actions, and environmental influences. SCT highlights three essential aspects of decision-making: internal motivations shaped by needs and context, planned actions influenced by attitudes and norms, and adaptive learning from social feedback. These principles naturally map to three key modules of our agent architecture: \textbf{motivation}, \textbf{action planning}, and \textbf{learning}.

Figure~\ref{fig:framework} depicts how these modules interact within a closed-loop structure. Motivations drive the formation of goals, which are translated into actions; actions leave observable traces in the environment, which in turn inform future decisions through learning. To ensure that each module reflects solid principles of human behavior rather than arbitrary assumptions, we further ground the design in established behavioral science theories. Motivation formation is modeled using \textit{Maslow’s hierarchy of needs}~\cite{maslow1943theory} to capture layered human needs; action planning is guided by the \textit{Theory of Planned Behavior}~\cite{ajzen1991theory} to explain how intentions and perceived control shape decisions; and learning is inspired by \textit{Social Learning Theory}~\cite{bandura1977social}, emphasizing adaptation through observation and feedback. This theoretical grounding allows agents to dynamically adjust their behavior in response to changing social contexts, improving the realism and reliability of social simulations.

The shared structure also enables a single agent design to operate coherently across multiple domains such as mobility, economy, social interaction, and cognition, in contrast to prior works that construct separate models for each individual task.
Together, these mechanisms establish a flexible yet theoretically grounded foundation for building social agents capable of realistic, context-sensitive adaptation. The following sections describe each module in detail and explain how they collectively enable scalable, theory-driven social simulations. Core prompts used in the proposal social agent can be found in Appendix~\ref{appendix:core_prompts}.

\subsection{Motivation: Maslow's Hierarchy of Needs}
\label{sec:needs_plan}
Maslow's Hierarchy of Needs describes the hierarchical needs of individuals, outlining the progression from basic physiological needs to higher-level needs, and ultimately to specific actions. Ranging from basic physiological needs, such as hunger and fatigue, to advanced needs, like developing interpersonal relationships, the theory posits that fundamental needs must be met before more advanced needs are satisfied, supporting the prediction of an individual's behavior based on the fulfillment of specific needs. To equip these agents with the capability of autonomously generating their behaviors based on their needs, we integrate Maslow's hierarchy to enhance the realism of decision-making by designing an autonomous workflow that helps agents determine and adjust their needs.

For physiological needs, such as hunger and fatigue, which tend to evolve independently of subjective evaluation, they can be described by mathematical models. Following the classic drive-reduction theory~\cite{hull1945principles}, we use the following function to determine the value of the needs for a specific agent at time $t$:
\[
N^i_t = min\{N^i_{max}, N^i_{t-1} + \mathrm{Inc}^i_t - \Delta\mathrm{N}^i_{t-1}\}
\]
Here, \(\mathrm{Inc}^i_t\) denotes the natural increase in the need, and \(\Delta \mathrm{N}^i_{t-1}\) is the quantity of need satisfied by the agent's action in the previous time step. This model captures the accumulation of unmet needs over time, counteracted by effective behavioral fulfillment.

Compared to objective physiological needs, higher-level needs, such as safety and social belonging, are more subjective. Human autonomy can substantially influence the perceived satisfaction of these needs, thereby opening a design space for leveraging LLM-based perception and reasoning modules. 
Following the Affective Events Theory~\cite{weiss2005reflections}, which states the relationship between discrete events and individual cognition, we adopt an event-driven formulation to capture the evolution of subjective needs. Events are divided into passive events triggered by the environment (e.g., receiving negative feedback in a group setting) and active events initiated by the individual (e.g., seeking friends or help). Changes are grounded in an agent’s perception of both passive and active events:
\[
N^i_t = \min\Bigl\{N^i_{\max},\;
N^i_{t-1}
+ \sum_{k=1}^{K} \Delta N^i_{\mathrm{act}}(e_k)
+ \sum_{l=1}^{L} \Delta N^i_{\mathrm{pas}}(e_l)
\Bigr\}
\]
where \(\Delta N^i_{\mathrm{act}}(e_k)\) represents the increment due to the \(k\)-th active event; \(\Delta N^i_{\mathrm{pas}}(e_l)\) represents the increment due to the \(l\)-th passive event; \(K\) is the total number of active events at time \(t\); and \(L\) is the total number of passive events at time \(t\).
To determine the impact of a particular event on an agent’s needs, we employ LLM to serve as the agent’s cognitive layer for event appraisal and feedback synthesis. Using the chain-of-thought (CoT) approach, the agent integrates information from its profile, current state, past experiences, and the specifics of the event to infer the subjective impact of the event on internal needs.

The model specifies the conditions and processes governing a social agent’s need dynamics, thereby supplying the motivational basis for its concrete actions. In the ensuing section, we examine how this motivation is converted into a concrete, executable sequence of actions.

\subsection{Scheduled Action: Theory of Planned Behavior}
\label{sec:multi_behavior}
The Theory of Planned Behavior (TPB) \cite{ajzen1991theory} posits that an individual’s behavior is driven by behavioral intentions, which are shaped by three cognitive factors: attitude toward the behavior, perceived social norms, and perceived behavioral control. In the context of social agents, TPB provides a structured mechanism for evaluating and selecting actions based on internal preferences, social expectations, and situational feasibility. Building on the motivational states established in Section~\ref{sec:needs_plan}, we adopt TPB to guide the transition from needs to actionable behavioral intentions. Rather than choosing actions arbitrarily, agents use TPB-aligned reasoning to compare alternative behavior options and select the one most consistent with their goals and social context.

\begin{table*}[ht]
\centering
\small
\begin{tabular}{c c c c c}
\toprule
\textbf{Module} & \textbf{Behavioral Property} & \textbf{Scenario} & \textbf{Comparison} & \textbf{Dataset} \\
\midrule
Motivation & Behavior varies in response to internal needs & Mobility \& Cognition & Full vs -woM & Beijing Mobility\\
Planning & Behavior reflects tradeoffs among contexts & Social \& Mobility & Full vs -woP & NYC Foursquare \\
Learning & Behavior adapts with accumulated experience & Economy \& Adaptation & Full vs -woL & SafeGraph + CBG\\
\bottomrule
\end{tabular}
\caption{Mapping between cognitive modules, behavioral properties, and evaluation settings.}
\label{tab:evaluation_mapping}
\end{table*}

\begin{figure}[tbp]
  \centering
  \includegraphics[width=0.7\linewidth]{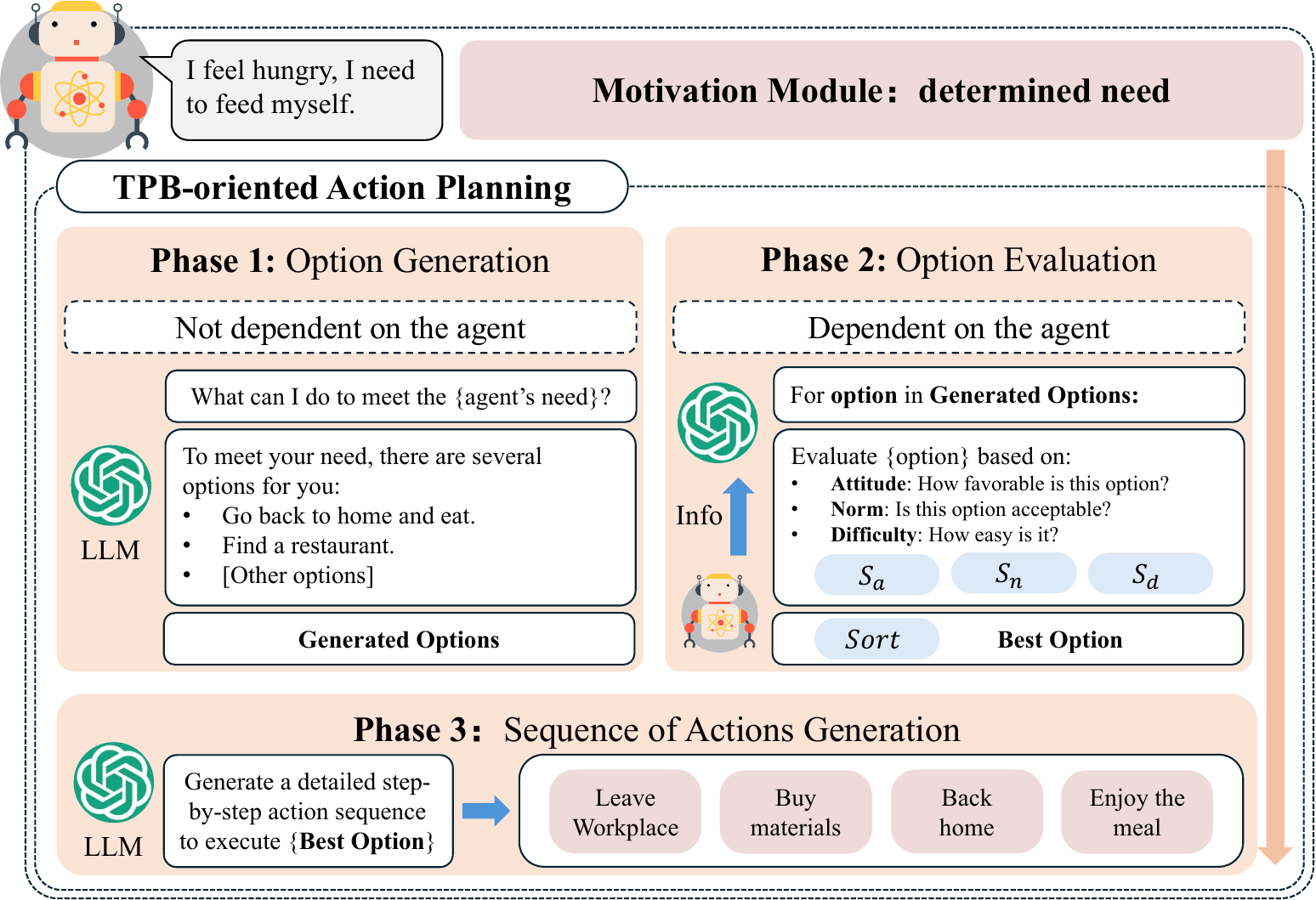}
  \caption{TPB-guided action generation workflow. }
  \label{fig:tpb_workflow}
\end{figure}

We implement a structured action generation workflow that connects motivational states to final behavioral execution. Given an activated need \( N_t \), the agent first generates a set of candidate behavioral goals \( \mathcal{A}_t = \{a_1, a_2, \dots, a_n\} \). Each candidate \( a_i \in \mathcal{A}_t \) is then evaluated according to the Theory of Planned Behavior (TPB), producing a corresponding intention score \( I(a_i) \). The final action is selected by:
\[
a^*_t = \arg\max_{a_i \in \mathcal{A}_t} I(a_i)
\]
The selected behavior \( a^*_t \) is subsequently grounded into an executable action sequence. This workflow formalizes how internal motivation is systematically transformed into interpretable and socially grounded behavior. Figure~\ref{fig:tpb_workflow} illustrates the overall process.

Each behavioral candidate \( a_i \) is evaluated along three cognitive dimensions defined by TPB: attitude (\( \mathrm{Att}(a_i) \)), subjective norm (\( \mathrm{Norm}(a_i) \)), and perceived difficulty (\( \mathrm{Ctrl}(a_i) \)). Respectively, they represent the agent’s internal preference, perceived social approval, and estimated feasibility of the action. The final intention score is computed as:
\[
I(a_i) = w_{\mathrm{att}} \cdot \mathrm{Att}(a_i) + w_{\mathrm{norm}} \cdot \mathrm{Norm}(a_i) + w_{\mathrm{ctrl}} \cdot \mathrm{Ctrl}(a_i)
\]
The agent infers all components through LLM-based reasoning over its context. This formulation provides an interpretable basis for comparing and selecting behavior options.

The TPB-guided reasoning process enables agents to evaluate behavior options in a cognitively structured manner. However, the quality and consistency of such evaluations fundamentally depend on the agent's access to relevant contextual knowledge—past experiences, social expectations, and situational cues. In the following section, we address how agents acquire, update, and internalize such information through interaction and observation, enabling continual adaptation over time.

\subsection{Learning and adaptation: Social Learning Theory}

\begin{figure}[t]
  \centering
  \includegraphics[width=0.7\linewidth]{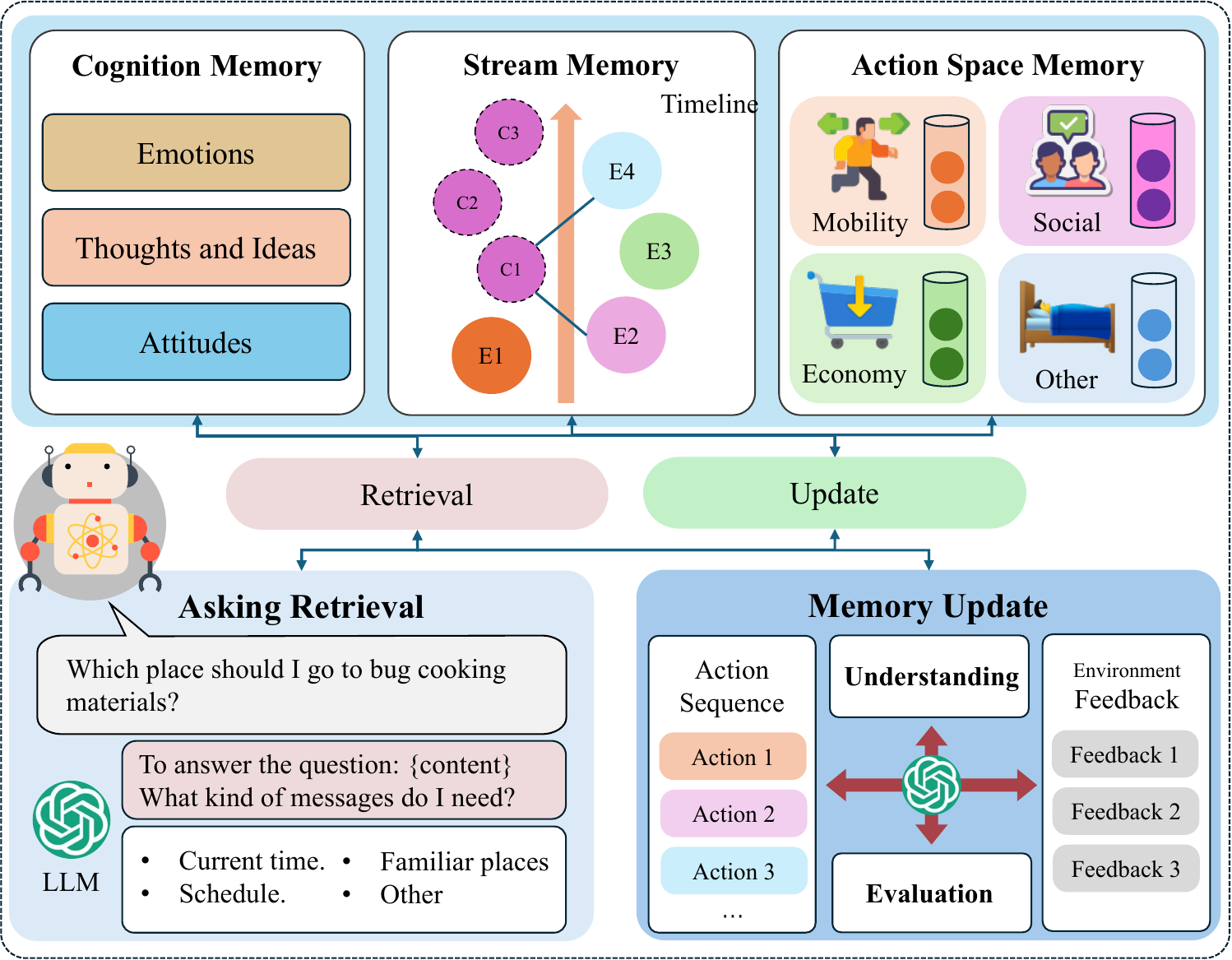}
  \caption{Learning mechanism of social agent}
  \label{fig:learning_framework}
\end{figure}

Social Learning Theory (SLT) \cite{bandura1977social} characterizes learning as a recursive process in which behavior is shaped through continuous interactions between internal cognition, experiential feedback, and environmental influence. 
As illustrated in Figure~\ref{fig:learning_framework}, in the context of LLM-based social agents, we abstract this learning process as a structured form of information processing. Specifically, we propose a two-part framework comprising: (1) a memory system that persistently stores, organizes, and links internal states, experiences, and actions; and (2) a retrieval mechanism that enables the agent to access relevant information through internal reasoning.

The memory system is composed of three functional components: \textbf{Stream Memory}, \textbf{Action Space Memory}, and \textbf{State Memory}. Stream Memory stores time-ordered records of the agent’s experiences and events; Action Space Memory organizes domain-specific behavioral experiences; and State Memory maintains evolving internal states, including emotions, attitudes, and contextual variables such as location.
Beyond storage, the memory system supports cognitive structuring through two mechanisms: (1) cross-modal association, allowing nodes such as attitudes to be linked with relevant events; and (2) abstraction, enabling the summarization of low-level experiences into higher-level representations. Together, these capabilities transform passive records into structured, retrievable knowledge.

To utilize stored memory, the agent employs a retrieval mechanism that mimics a human-like reasoning process. Instead of relying on surface-level similarity or predefined queries, the agent first examines the current situation and actively formulates a cognitive prompt: What information do I need to understand better or assess this context? This internally generated query then guides memory retrieval, allowing the agent to activate only the most relevant, situation-specific content.
By grounding retrieval in contextual reflection, Asking Retrieval transforms information access from a passive lookup process into an integrated component of the reasoning workflow. It aligns the memory system with the agent’s goals, perceptions, and current environment, enabling adaptive behavior and cognitively coherent decision-making.

The integration of structured memory and retrieval enables agents to accumulate experiences, abstract knowledge, and adapt behavior over time. This learning mechanism transforms past interactions into actionable insights, shaping not only future decisions but also the agent’s evolving attitudes, preferences, and perceived behavioral control.
Our framework closes the cognitive loop: learned experiences feed back into the formation of motivation and the generation of actions, enabling coherent and adaptive behavior across interactions.

\label{sec:agent_workflow}
\section{Evaluation} \label{section:evaluation}

\subsection{Evaluation strategy}

The proposed social agent incorporates three cognitive modules: motivation modeling, action planning, and learning-based adaptation. Rather than evaluating agents through specific task performance, we aim to answer the question of \textbf{whether these theory-driven modules contribute to meaningful and consistent behavioral patterns under complex conditions.} For each module, we identify a core behavioral property it is intended to shape, and select a scenario where such effects are observable and theoretically grounded. In each case, we compare the full social agent with an ablated version lacking the target module to isolate its impact.

\textbf{Motivation modeling.}  
Motivation shapes whether and why an agent initiates a behavior. In human contexts, actions such as movement are rarely arbitrary; they are often driven by internal needs. An effective motivation module should produce behavioral variability that reflects differences in internal state. We evaluate this property in a scenario focused on daily mobility. We compare the full social agent with a version lacking the motivation mechanism (-woM), observing whether the presence of need dynamics results in more human-like movement patterns. The environment is instantiated using a high-resolution mobility dataset from Beijing, comprising 159 individual-level movement trajectories over seven days.

\textbf{Action planning.}  
Planning governs how a behavior is executed given an intention. When individuals experience a social need, they can choose from various options. Selecting an appropriate option involves weighing multiple factors: expected outcomes, social norms, physical constraints, etc. To test this capacity, we use a scenario centered on social mobility. We compare the full model with a variant without the planning module (-woP), assessing whether planning leads to more context-sensitive and cost-effective behavior. We built the environment with Location-Based Social Network data in NYC~\cite{10.1145/3308558.3313635}.

\textbf{Learning-based adaptation.}  
Learning enables agents to refine future behavior based on experience. This includes adjusting attitudes, preferences, and strategies over time. Without learning, agents behave myopically; with learning, their decisions should exhibit continuity, evolution, and situational awareness. We examine this through a long-horizon simulation modeled after pandemic disruptions. We compare the full model with a variant lacking memory and adaptation (-woL), evaluating whether experience accumulation improves behavioral stability and responsiveness. The evaluation environment is initialized with SafeGraph\footnote{\url{https://www.safegraph.com/}} visit flows combined with U.S. Census Survey\footnote{\url{https://www.census.gov/}} data.

Table~\ref{tab:evaluation_mapping} summarizes the mapping between modules, behavioral properties, scenarios, and ablation baselines. Detailed dataset descriptions are provided in Appendix~\ref{appendix:datasets}.

\subsection{Effect of motivation modeling on realistic mobility behavior}

\begin{table}[htbp]
\begin{center}
\begin{tabular}{lccccc}
\toprule
\multicolumn{1}{c}{\bf Method}   & \multicolumn{1}{c}{\bf Radius} & \multicolumn{1}{c}{\bf Dayloc} & \multicolumn{1}{c}{\bf itdNum} & \multicolumn{1}{c}{\bf itdError}\\
\midrule
Timegeo    & 0.254 & 0.258 & 0.297 & 0.536\\
Movesim    & 0.233 & 0.051 & 0.154 & 0.904\\
Volunteer  & 0.455 & 0.049 & 0.318 & 0.804\\
DiffTraj   & 0.027 & 0.647 & 0.695 & 0.597\\
Act2Loc    & 0.024 & 0.042 & 0.131 & 0.391\\
\midrule
\textbf{\-woM}       & \textbf{0.371} & \textbf{0.433} & \textbf{0.171} & \textbf{0.392}\\
\textbf{Ours}        & \textbf{0.023} & \textbf{0.038} & \textbf{0.073} & \textbf{0.094}\\
\bottomrule
\end{tabular}
\end{center}
\caption{The results of Mobility \& Cognition among LLM agents.}
\label{tab:exp1:performance}
\end{table}

This experiment examines whether integrating a motivation module enables agents to reproduce human-like mobility patterns. The evaluation is conducted along two complementary dimensions. The first dimension focuses on the mobility outcome of agents, captured by the radius of gyration, which reflects the spatial dispersion of movement, and the daily visited locations, which quantify the diversity of places visited. The second dimension focuses on the cognitive stimuli behind mobility, evaluated through the number of intentions generated per day and the similarity between intention sequences, both measuring the stability and coherence of decision-making processes. Table~\ref{tab:exp1:performance} summarizes the quantitative results across these metrics.

Our approach achieves the lowest errors on all metrics compared with classical generative models, including TimeGeo~\cite{timegeo}, Movesim~\cite{movesim}, Volunteer~\cite{volunteer}, DiffTraj~\cite{difftraj}, and Act2Loc~\cite{liu2024act2loc}, and the ablated version without the motivation module (woM). The absence of motivation modeling in woM leads to unstable and fragmented intention patterns, which further propagate to mobility behavior, resulting in inflated errors in spatial dispersion and daily location counts. In contrast, our method produces consistent and structured intentions that align with plausible human needs, yielding realistic and coherent movement patterns.

\begin{figure}[t]
\centering
\includegraphics[width=0.6\textwidth]{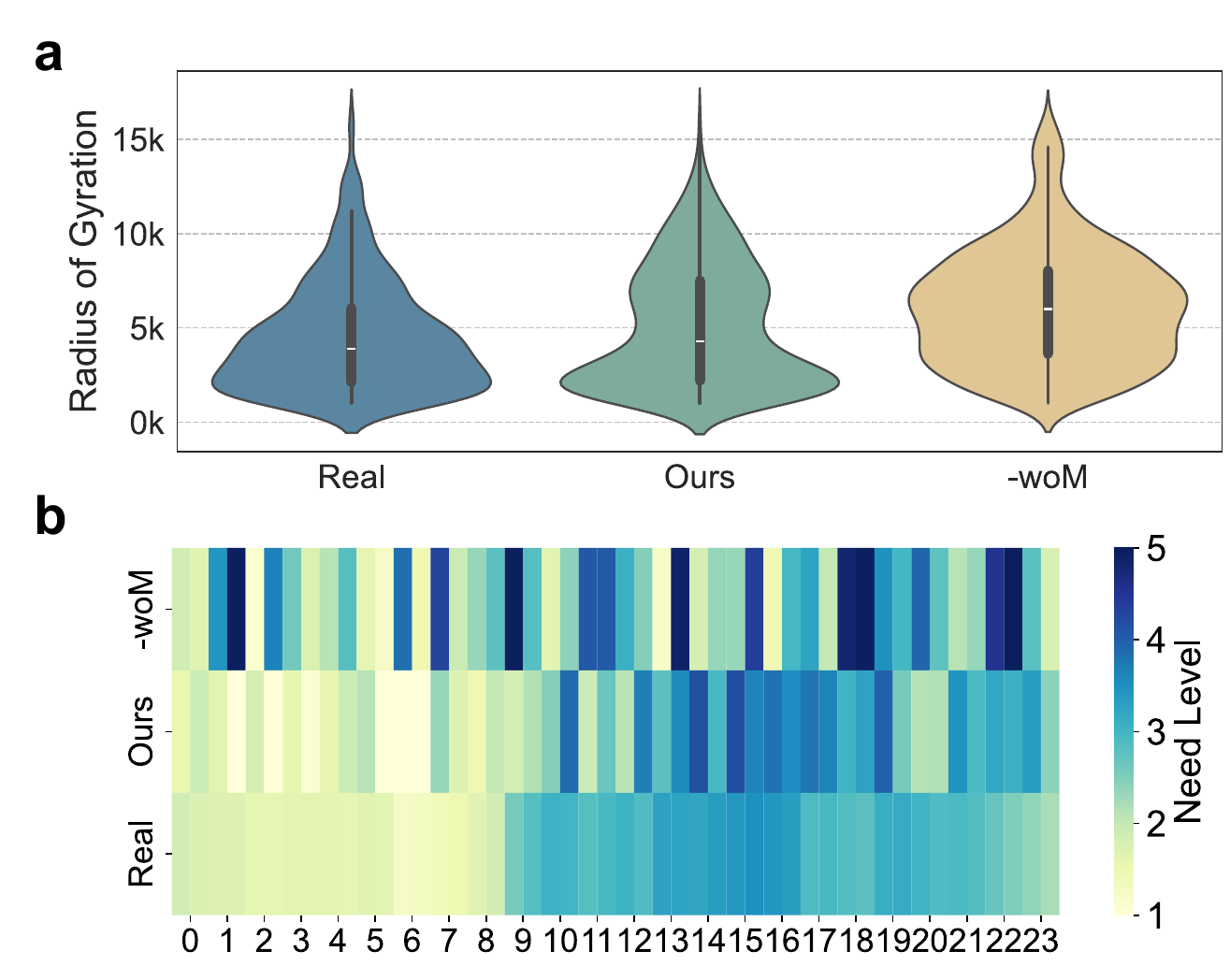}
\caption{Comparison of mobility and deed dynamics across methods. (a) Distribution of Radius of Gyration; (b) Temporal heatmap of Need Intensity over 48 time intervals.}
\label{fig:exp1:results}
\end{figure}

Figure~\ref{fig:exp1:results} provides a visual comparison of two representative metrics. Figure~\ref{fig:exp1:results} (a) shows the distribution of the mobility radius, where our method closely follows the long-tailed pattern of real human movements, while woM underestimates both short-range stays and long-distance travel. Figure~\ref{fig:exp1:results} (b) illustrates the temporal evolution of needs inferred from the generated intention sequences. Needs are categorized into five levels, ranging from physiological demands such as fatigue and hunger to safety and social interactions, following Maslow's hierarchy of needs. Our model reproduces stable, time-dependent peaks consistent with real-world patterns, whereas woM generates irregular fluctuations without clear behavioral rhythms.

In general, these results demonstrate that realistic mobility cannot be achieved by movement rules alone. Explicitly modeling motivation stabilizes the generation of behavioral intentions, which in turn leads to spatially and temporally coherent mobility. The strong coupling between needs and actions highlights the necessity of motivation-driven design in constructing socially aligned LLM agents.

\subsection{Contribution of action planning to realistic social interaction decisions}
\begin{table}[h]
\centering
\begin{tabular}{cccc}
\toprule
\textbf{Method} & \textbf{Proportion (\%)} & \textbf{Median (km)} & \textbf{RelError (\%)} \\
\midrule
Real    & 15.4 & 5.02 & -- \\
woP    & 23.3 & 8.06 & 60.4 \\
Ours    & 12.3 & 3.44 & \textbf{31.6} \\
\bottomrule
\end{tabular}
\caption{Statistics of social trips under different settings. Relative error is calculated against real-world median distances.}
\label{tab:exp2:stats}
\end{table}
This experiment evaluates how the action planning module contributes to realistic decision-making in social interactions. Both our approach and the ablation model without planning are driven by the same level of social needs, ensuring that differences in outcomes are attributed solely to planning. The planning module is responsible for two key decisions: first, whether an offline meeting is necessary compared to online communication, and second, which target to visit when multiple options exist. These decisions are crucial for avoiding unrealistic or excessive travel behavior.

Table~\ref{tab:exp2:stats} reports key statistics on social trips, including their proportion among all trips and their travel distance distribution. In real-world data, 15.4\% of trips are dedicated to social visits. Our approach generates 12.3\%, closely matching this natural frequency, while the model without planning produces an unrealistic 23.3\%, overestimating social trips by more than 50\%. This demonstrates that, without the support of the planning module, agents fail to make informed decisions between offline and online social interactions, leading to an unrealistic overemphasis on offline visits.

\begin{figure}[t]
\centering
\includegraphics[width=0.6\textwidth]{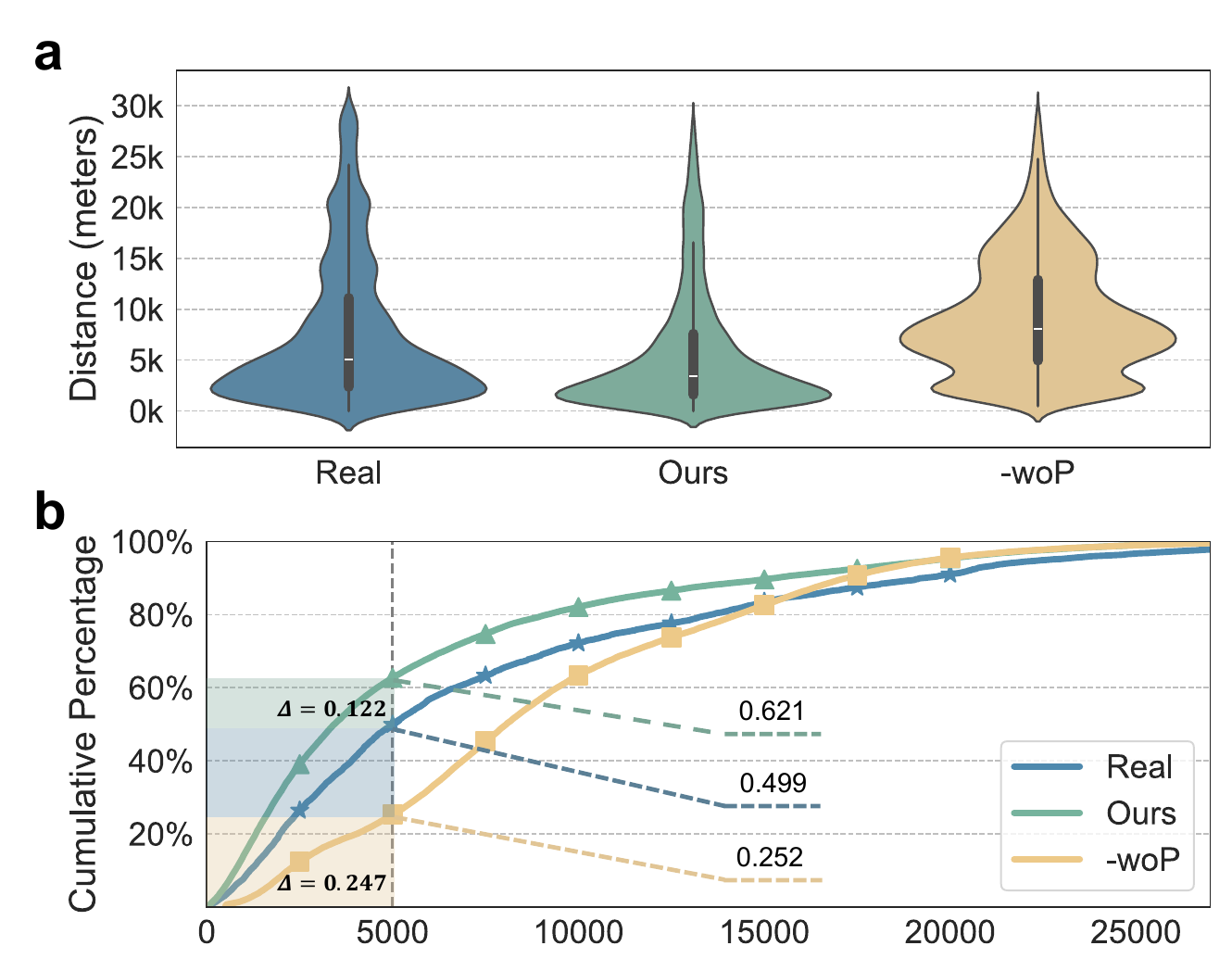}
\caption{Comparison of social trips across methods: (a) Trip distance distribution; (b) Cumulative visit ratio by distance.}
\label{fig:exp2:results}
\end{figure}

For travel distances, our method keeps social trips localized, with a median distance of 3.44 km and a relative error of 31.6\% compared to real data. The non-planning baseline produces much longer trips, with a median of 8.06 km and a 60.4\% deviation. Figure~\ref{fig:exp2:results} provides a finer-grained view of this pattern. By examining the cumulative distribution, almost half of real-world social trips (49.9\%) occur within 5 km. Our approach captures this local tendency even more strongly (62.1\%), whereas the absence of planning results in only 25.2\% of social trips within the same range. The lack of planning thus leads to an unrealistic dominance of long-distance trips.

Overall, the results highlight that action planning regulates not only the frequency of offline social interactions but also their spatial scale. By enabling rational decision-making on whether, whom, and how far to visit, planning ensures that social mobility behaviors remain aligned with real-world patterns.

\subsection{Impact of learning mechanism on long-term behavioral adaptation during pandemic}

This experiment aims to verify the effectiveness of the proposed learning module in enabling social agents to adapt their behavior over time. We select a pandemic scenario where mobility patterns experience a sudden drop due to the introduction of movement restrictions. This setup provides a clear testbed to evaluate whether agents can gradually adjust their mobility in response to environmental changes.

\begin{figure}[t]
\centering
\includegraphics[width=0.6\textwidth]{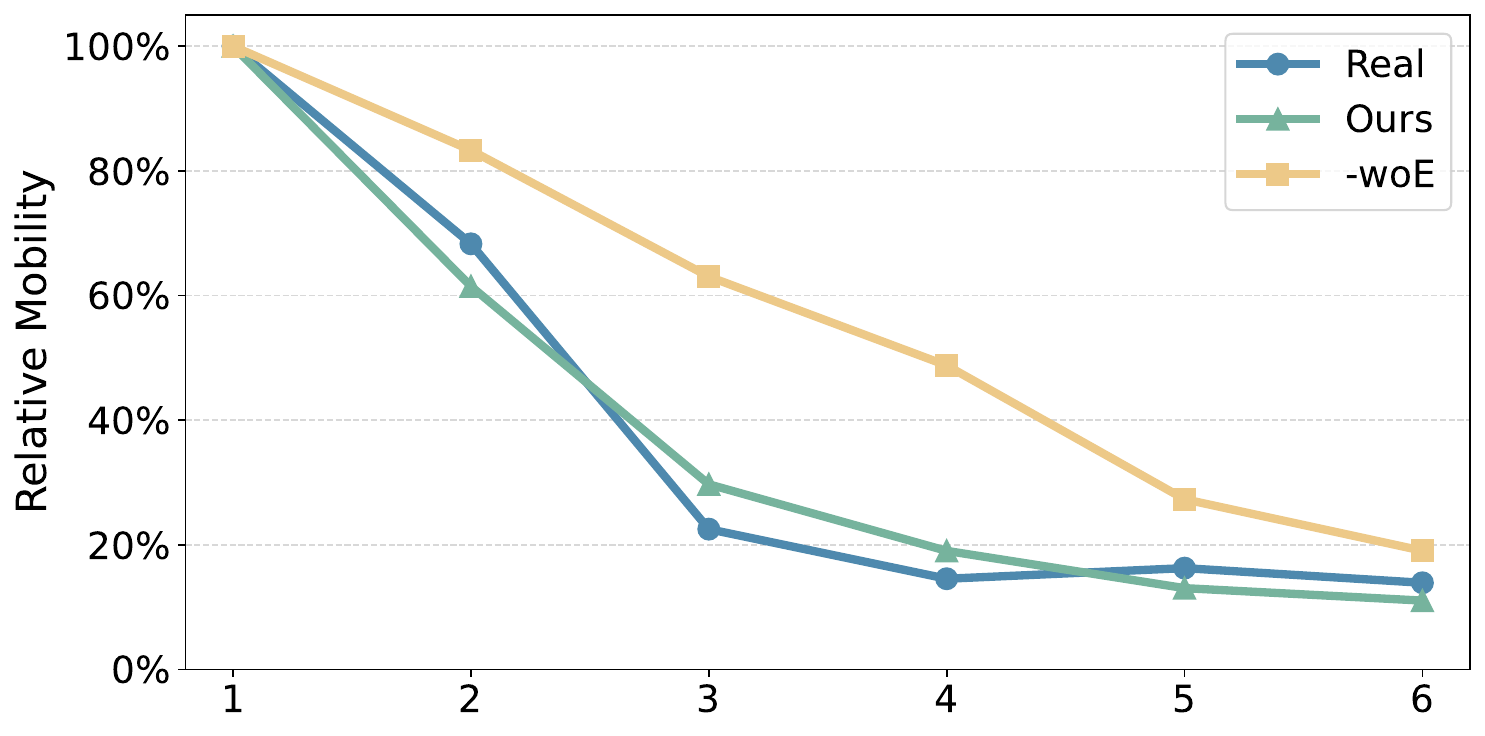}
\caption{Comparative temporal trends of relative mobility across real-world data, our method, and ablation (-woE)}
\label{fig:exp3:mobility_trend}
\end{figure}

Figure~\ref{fig:exp3:mobility_trend} shows the relative mobility trend across six weeks. Our method closely tracks real mobility changes, capturing both the sharp drop after restrictions and the gradual stabilization. For example, in week 3, our model maintains 0.297 mobility compared to the real 0.225, while -woE remains much higher at 0.630. Similarly, by week 6, our approach reaches 0.110 versus the real 0.139, whereas -woE stays higher at 0.190. The variant without the learning module (-woE) exhibits a near-linear decline, failing to capture the slowing decrease and rebound trend observed in real data. Although our approach shows better adaptation, it slightly underestimates mobility in the later weeks, suggesting that LLM agents are more sensitive to environmental changes compared to real individuals.

\begin{figure}[t]
\centering
\includegraphics[width=0.6\textwidth]{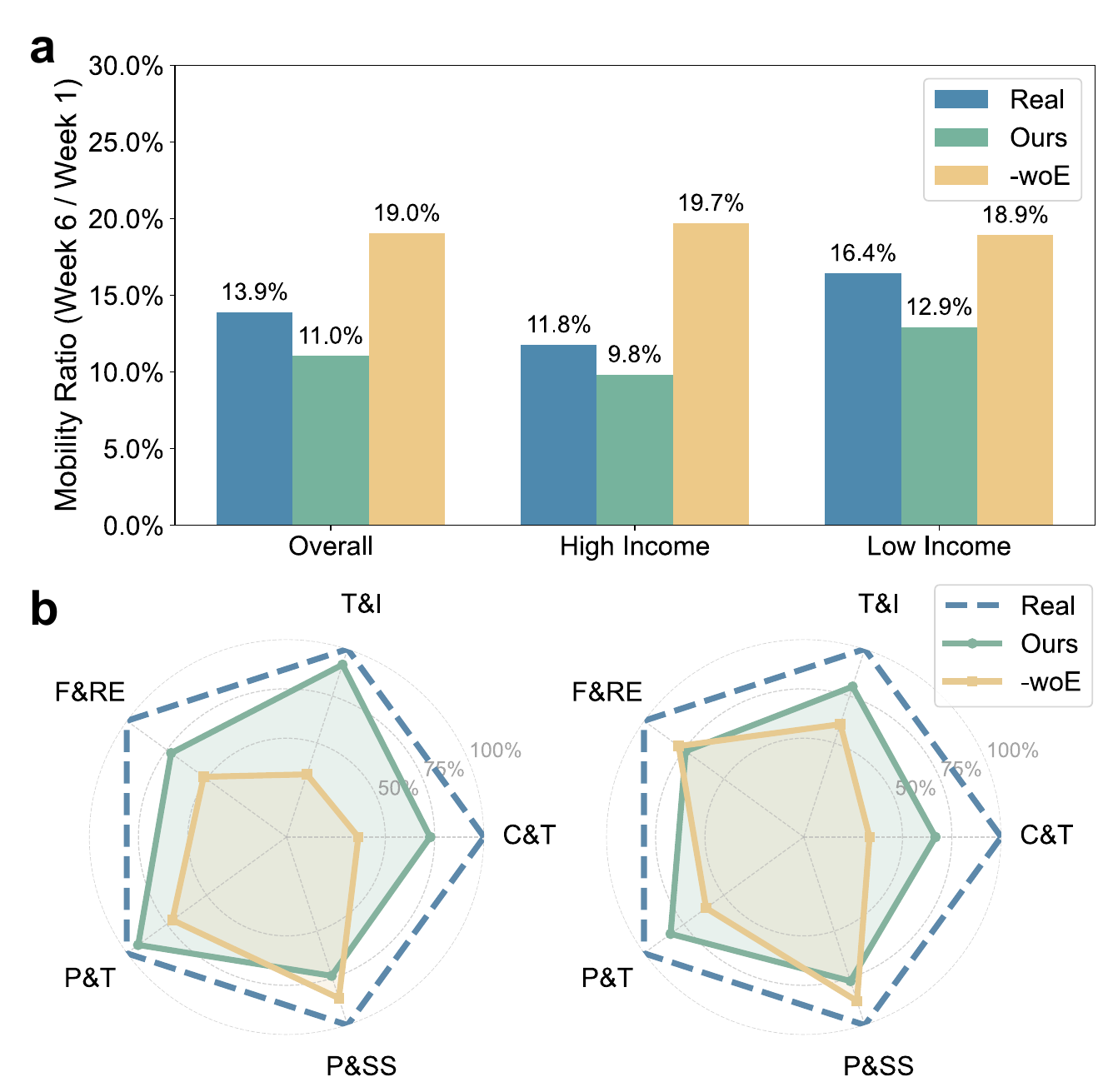}
\caption{Assessment of mobility ratio changes and poi visitation patterns across income groups.}
\label{fig:exp3:income_dynamics}
\end{figure}

Figure~\ref{fig:exp3:income_dynamics} evaluates whether social agents can reproduce behavioral differences across income groups. Figure~\ref{fig:exp3:income_dynamics}(a) shows the overall mobility ratio reduction from week~6 to week~1. Our method preserves the relative gap between high-income and low-income groups, achieving reductions of $9.8\%$ and $12.9\%$, close to the real-world values of $11.8\%$ and $16.4\%$. In contrast, -woE produces nearly uniform reductions of around $19\%$ for both groups, failing to reflect population heterogeneity. Figure~\ref{fig:exp3:income_dynamics}(b) further analyzes the similarity of POI visitation patterns compared to real data. Our approach achieves consistently higher alignment scores across most categories; for example, in high-income groups, the alignment for C\&T (Commerce and Trade) is $0.73$ compared to $0.36$ for woE, and in low-income groups, the score for P\&T (Professional \& Technical) is $0.83$ compared to $0.61$. These results demonstrate that the learning module enables agents to internalize differentiated adaptation strategies for distinct populations, whereas agents without learning converge to nearly homogeneous and less realistic behaviors regardless of income level.

Overall, the results highlight the necessity of the learning module for enabling realistic long-term behavioral adaptation. Without it, agents fail to adjust their behaviors dynamically in response to evolving external conditions, leading to less plausible and less human-like simulation outcomes.

\section{Conclusion}

Inspired by Social Cognition Theory, this work introduces a theory-driven design for constructing generative social agents powered by large language models. The proposed design integrates three key modules: motivation modeling, action planning, and learning, endowing agents with cognitively grounded decision-making capabilities. Comprehensive experiments across multiple scenarios demonstrate that each module substantially enhances the overall effectiveness and reliability of the proposed social agent. These findings highlight the value of embedding established social science theories into LLM-based agent architectures, paving the way toward more realistic, interpretable, and reliable social simulations.

\clearpage

\bibliographystyle{unsrt}  
\bibliography{references}

\clearpage

\appendix
\section{Core Prompts used in Social Agent} \label{appendix:core_prompts}

This appendix presents the core prompts used for the three cognitive modules of the social agent: \textit{Motivation}, \textit{Action Planning}, and \textit{Learning}. We first clarify how various types of information are referenced within these prompts to ensure clarity and consistency.

\subsection{Information Referencing Conventions}

The prompts incorporate three main categories of information:

\begin{itemize}
  \item \textbf{Agent Profile:} Static and dynamic attributes such as demographics, personality, preferences, and current internal states. Variables are referenced using \texttt{\{variable\_name\}} syntax, e.g., \texttt{\{age\}}, \texttt{\{current\_emotion\}}.

  \item \textbf{Memory Information:} Past experiences and learned knowledge stored in memory. Queries to memory use the format:  
  \begin{quote}
    \texttt{[Memory Query]: <natural language question>}
  \end{quote}
  with results inserted as:  
  \begin{quote}
    \texttt{[Memory Retrieval Result]: <retrieved content>}
  \end{quote}

  \item \textbf{Environment Context:} External factors including time, weather, location, nearby POIs, and social events, presented as structured text blocks with variables indicated by \texttt{\{variable\_name\}}.
\end{itemize}

Each prompt clearly labels these information blocks to guide the language model’s reasoning.

\subsection{Motivation Module Prompts}

The motivation module models the agent's internal needs and their dynamics.

\textbf{Prompt 1: Initialization of Basic Needs}

This prompt initializes the agent’s basic physiological needs such as hunger and fatigue by integrating static profile attributes and current environmental conditions.

\begin{tcolorbox}[promptbox,title={Prompt 1: Initialization of Basic Needs}]
Given the agent profile: \\
Name: \{name\} \\
Age: \{age\} \\
Health Status: \{health\_status\} \\

And current environment: \\
Time: \{current\_time\} \\
Weather: \{weather\} \\

Estimate the agent’s initial levels of hunger and fatigue based on these factors. Provide a structured summary with values from 0 (none) to 1 (maximum).

Example Output:

\{
  "hunger": 0.4, \\
  "fatigue": 0.2 \\
\}
\end{tcolorbox}

\vspace{0.5cm}
\textbf{Prompt 2: Initialization of High-Level Needs}

This prompt estimates high-level needs, such as social belonging, by integrating the agent’s profile, memory-based social network information, and environmental context.

\begin{tcolorbox}[promptbox,title={Prompt 2: Initialization of High-Level Needs}]
Agent Profile: \{profile\} \\

[Memory Query]: What recent social interactions has the agent engaged in? What is the current status of the agent’s social network? \\

[Memory Retrieval Result]: \{retrieved\_social\_memory\} \\

Based on this information and the agent’s profile, estimate the current social need level with reasoning. Output a score between 0 and 1.

Example Output:

\{\\
  "social\_need": 0.75, \\
  "reasoning": "Agent has limited recent social interactions but active social network presence." \\
\}
\end{tcolorbox}

\vspace{0.5cm}
\textbf{Prompt 3: Dynamic Update of Needs}

This prompt updates the agent’s needs dynamically in response to recent active and passive events, leveraging memory retrieval to inform the update.

\begin{tcolorbox}[promptbox,title={Prompt 3: Dynamic Update of Needs}]
Agent's current needs: \{current\_needs\} \\

Recent active events: \{active\_events\} \\
Recent passive events: \{passive\_events\} \\

[Memory Query]: How have similar events affected the agent’s needs previously? \\

[Memory Retrieval Result]: \{retrieved\_memory\} \\

Provide step-by-step reasoning and updated needs.

Example Output:

\{\\
  "updated\_needs": \{ \\
    \quad "hunger": 0.5, \\
    \quad "fatigue": 0.3, \\
    \quad "social\_need": 0.8 \\
  \}, \\
  "reasoning": "Recent social rejection increased social need; physical activity reduced fatigue." \\
\}
\end{tcolorbox}

\subsection{Action Planning Module Prompts}

The action planning module transforms needs into concrete behavior plans.

\textbf{Prompt 1: Generate Candidate Actions}

This prompt instructs the model to produce a diverse list of possible actions to satisfy the agent’s current activated need.

\begin{tcolorbox}[promptbox,title={Prompt 1: Generate Candidate Actions}]
Agent’s current need: \{need\_description\} \\

Based on the profile and environment, generate possible actions to satisfy this need.

Example Output:

[\\
  "Go home and cook dinner", \\
  "Order food delivery", \\
  "Visit a nearby restaurant" \\
]
\end{tcolorbox}

\vspace{0.5cm}
\textbf{Prompt 2: Score Candidate Actions Using TPB and Memory}

This prompt evaluates each candidate action’s attitude, subjective norm, and perceived control, integrating relevant past experiences from memory.

\begin{tcolorbox}[promptbox,title={Prompt 2: Score Candidate Actions Using TPB and Memory}]
For each candidate action: \{action\_description\} \\

[Memory Query]: What is the agent’s past attitude and experience with this action? \\

[Memory Retrieval Result]: \{retrieved\_memory\} \\

Using profile, social norms, and memory, score actions on: \\
- Attitude (0–1) \\
- Subjective Norm (0–1) \\
- Perceived Behavioral Control (0–1) \\

Provide detailed reasoning for each score.

Example Output:

[\\
  \{ \\
    "action": "Go home and cook dinner", \\
    "attitude": 0.9, \\
    "subjective\_norm": 0.8, \\
    "perceived\_control": 0.7 \\
  \}, \\
  \{ \\
    "action": "Order food delivery", \\
    "attitude": 0.5, \\
    "subjective\_norm": 0.6, \\
    "perceived\_control": 0.9 \\
  \}, \\
  \{ \\
    "action": "Visit a nearby restaurant", \\
    "attitude": 0.6, \\
    "subjective\_norm": 0.7, \\
    "perceived\_control": 0.5 \\
  \} \\
]
\end{tcolorbox}

\vspace{0.5cm}
\textbf{Prompt 3: Generate Detailed Action Sequence}

This prompt generates a step-by-step plan describing how the agent will execute the best-scored action.

\begin{tcolorbox}[promptbox,title={Prompt 3: Generate Detailed Action Sequence}]
Best candidate action: \{best\_action\} \\

Generate a detailed action sequence describing how the agent will execute this action.

Example Output:

[\\
  "Finish current work tasks", \\
  "Leave office", \\
  "Go to grocery store to buy ingredients", \\
  "Return home", \\
  "Cook and eat dinner" \\
]
\end{tcolorbox}

\subsection{Learning Module Prompts}

The learning module enables adaptation through reflection on experience and memory.

\textbf{Prompt 1: Generate Agent Thoughts for an Event}

This prompt produces the agent’s internal thoughts, attitudes, and reflections about a specific event.

\begin{tcolorbox}[promptbox,title={Prompt 1: Generate Agent Thoughts for an Event}]
Agent Profile: \{profile\} \\
Event: \{event\_description\} \\

[Memory Query]: What relevant past experiences and attitudes relate to this event? \\

[Memory Retrieval Result]: \{retrieved\_memory\} \\

Generate the agent’s thoughts, attitudes, and reflections about this event. \\
Example Output:

\{\\
  "thoughts": "I feel disappointed by the cancellation but understand the reasons.", \\
  "attitude": "Negative towards last-minute changes.", \\
  "reflection": "I should prepare backup plans in future." \\
\}
\end{tcolorbox}

\vspace{0.5cm}
\textbf{Prompt 2: Update Emotional State}

This prompt updates the agent’s emotional state based on recent events and retrieved memory.

\begin{tcolorbox}[promptbox,title={Prompt 2: Update Emotional State}]
Current emotion: \{current\_emotion\} \\

Recent events: \{recent\_events\} \\

[Memory Query]: How have similar events affected emotions before? \\

[Memory Retrieval Result]: \{retrieved\_memory\} \\

Update the emotional state with reasoning.

Example Output:

\{\\
  "updated\_emotion": "frustrated", \\
  "reasoning": "Repeated delays in plans cause increased frustration." \\
\}
\end{tcolorbox}

\vspace{0.5cm}
\textbf{Prompt 3: Structure Recent Experiences for Memory}

This prompt summarizes recent experiences, emotions, and responses into structured memory entries.

\begin{tcolorbox}[promptbox,title={Prompt 3: Structure Recent Experiences for Memory}]
Summarize recent events, emotions, and responses into structured entries for memory.

Example Output:

[\\
  \{ \\
    "event": "Visited restaurant", \\
    "emotion": "satisfied", \\
    "outcome": "hunger reduced" \\
  \}, \\
  \{ \\
    "event": "Received negative social feedback", \\
    "emotion": "disappointed", \\
    "outcome": "increased social need" \\
  \} \\
]
\end{tcolorbox}

\vspace{0.5cm}
\textbf{Prompt 4: Formulate Memory Retrieval Queries}

This prompt generates retrieval queries based on the current context to assist decision-making.

\begin{tcolorbox}[promptbox,title={Prompt 4: Formulate Memory Retrieval Queries}]
Current context: \{context\} \\

Generate questions to retrieve relevant past experiences.

Example Output:

[\\
  "How did I react to similar weather conditions?", \\
  "What actions did I take after feeling fatigued?", \\
  "What social activities improved my mood previously?" \\
]
\end{tcolorbox}

\vspace{0.5cm}
\textbf{Prompt 5: Abstract General Behavioral Strategies}

This prompt extracts generalized strategies from accumulated memories for guiding future behavior.

\begin{tcolorbox}[promptbox,title={Prompt 5: Abstract General Behavioral Strategies}]
From accumulated memories, abstract generalized strategies for future behaviors.

Example Output:

\{\\
  "strategy\_1": "Prefer short trips when moderately hungry.", \\
  "strategy\_2": "Avoid outdoor social activities during bad weather.", \\
  "strategy\_3": "Seek social support when feeling isolated." \\
\}
\end{tcolorbox}

\section{Evaluation Datasets} \label{appendix:datasets}
To support the evaluation described above, we construct a composite simulation environment grounded in real-world data. These datasets are selected to ensure that the agent operates under realistic conditions, allowing for direct behavioral comparison with human traces.

\textbf{Mobility \& Cognition.}
We utilize a high-resolution mobility dataset from Beijing, comprising 159 individual-level movement trajectories collected over a one-week period. Each record contains location timestamps and semantic POI types. 
This dataset enables us to evaluate whether the agent's movement patterns are influenced by internal need states and whether these decisions exhibit human-like rhythms, such as commuting or need-based detours.

\textbf{Social \& Mobility.}
This experiment utilizes a two-week check-in dataset from Foursquare in New York City, capturing individual-level visits to venues along with social tags and user activity profiles. 
It enables us to examine how agents with social needs choose physical activities, and whether planned behavior aligns with social context and intention strength.

\textbf{Economy \& Adaptation.}
We adopt data from Safegraph and U.S. Census Block Groups (CBG) to simulate the impact of the COVID-19 pandemic on mobility under economic constraints. Safegraph provides POI visit volumes and movement flows, while CBG provides demographic and income-level profiles. 
This long-horizon setting allows us to observe whether agents adapt over time by adjusting behavior in response to dynamic environmental changes and economic pressures.

All scenarios are simulated using the AgentSociety Platform, with multiple agents instantiated from human-like profiles. The underlying simulation environment supports synchronized time steps, environment feedback, and interaction logging.

\end{document}